\documentclass{bvm2022}

\begin{document}

\selectlanguage{english}

\title{Comparison of Depth Estimation Setups from Stereo Endoscopy and Optical Tracking for Point Measurements}

\titlerunning{Comparison of Depth Estimation Setups}

\author{
	Lukas \lname{Burger} \inst{1}, 
	Lalith \lname{Sharan} \inst{1,4}, 
	Samantha \lname{Fischer} \inst{1}, 
	Julian \lname{Brand} \inst{1}, 
	Maximillian \lname{Hehl} \inst{1}, 
	Gabriele \lname{Romano} \inst{2},
	Matthias \lname{Karck} \inst{2},
	Raffaele \lname{De~Simone} \inst{2},
	Ivo \lname{Wolf} \inst{3},
	Sandy \lname{Engelhardt} \inst{1,4}
}

\authorrunning{Burger et al.}

\institute{
\inst{1} Group Artificial Intelligence in Cardiovascular Medicine (AICM),\\
Department of Internal Medicine III, Heidelberg University Hospital, Heidelberg\\
\inst{2} Department of Cardiac Surgery, Heidelberg University Hospital, Heidelberg\\
\inst{3} Mannheim University of Applied Sciences, Mannheim\\
\inst{4} DZHK (German Centre for Cardiovascular Research), partner site Heidelberg/Mannheim
}

\email{sandy.engelhardt@med.uni-heidelberg.de}

\maketitle

\begin{abstract}
To support minimally-invasive intraoperative mitral valve repair, quantitative measurements from the valve can be obtained using an infra-red tracked stylus. It is desirable to view such manually measured points together with the endoscopic image for further assistance. Therefore, hand-eye calibration is required that links both coordinate systems and is a prerequisite to project the points onto the image plane. 
A complementary approach to this is to use a vision-based endoscopic stereo-setup to detect and triangulate points of interest, to obtain the 3D coordinates. In this paper, we aim to compare both approaches on a rigid phantom and two patient-individual silicone replica which resemble the intraoperative scenario. 
The preliminary results indicate that 3D landmark estimation, either labeled manually or through partly automated detection with a deep learning approach, provides more accurate triangulated depth measurements when performed with a tailored image-based method than with stylus measurements.

\end{abstract}

\section{Introduction}

In mitral valve repair (MVR), sutures are placed around the mitral valve annulus, and a ring prosthesis is placed through the sutures on the annulus, to perform \textit{annuloplasty}. MVR is increasingly performed in a minimally invasive setup \cite{3286-casselman_filip_p._mitral_2003}, enabled through endoscopic video display. 
In this context, $3D$ endoscopes are used more frequently since they facilitate better depth perception.
Furthermore, patient-specific physical simulators have been developed for the purpose of surgical training  and pre-operative planning \cite{3286-DBLP:conf/miigp/Boone0GBEP19,3286-EngelhardtIJCARS2019}.
During surgery, the valvular pathomorphology is traditionally assessed by visual intra-operative exploration \textit{in-situ} \cite{3286-engelhardt_accuracy_2016}.
However, this approach lacks a quantitative base and is therefore not easily comparable between surgeons. In order to increase reproducibility and for intraoperative decision support, Engelhardt et al. \cite{3286-engelhardt_accuracy_2016} proposed the use of an assistance system for measuring dimensions of the mitral valve geometry (e.g. length of the chordae tendineae, width and shape of the annulus). The system incorporates infra-red based optical stereo tracking of customized instruments that are equipped with spherical markers. It has been successfully applied in $9$ patients during surgery \cite{3286-Engelhardt2016ATS}. 
A systematic accuracy investigation  \cite{3286-engelhardt_accuracy_2016} revealed that 
phantom experiments conducted in the actual application environment (OR) deliver a high system accuracy (mean precision $0.12 \pm 0.093\ts \text{mm}$, mean trueness $0.77 \pm 0.39\ts \text{mm}$) and a low user
error (mean precision $0.18 \pm 0.10\ts \text{mm}$, mean trueness
$0.81 \pm 0.36\ts \text{mm}$).
However, it is cumbersome to setup the system before the surgery. Moreover, the measurements are performed manually, meaning that the user needs to be experienced with the system.
Furthermore, in the current setup, the information stream gained from the optical tracking system is not registered to the endoscopic system, which was criticized by the end users. They prefer to have the measurements displayed in the endoscopic image as a \elqq quality check\erqq{} potentially together with additional information using augmented reality \cite{3286-EngelhardtMiccai14}.

In this work, we perform a hand-eye calibration of the stereo-endoscopic camera to display the measured points in the endoscopic frames. This introduces additional sources of error caused by hand-eye calibration, which includes estimation of intrinsic and extrinsic camera calibration parameters. 
We investigate the errors produced by this setup on a rigid phantom and two patient-specific silicone valves mounted on a simulator. Since errors in systems that include an optical tracking device also relate to the individual instrument design, they need to be assessed in an application-related setup \cite{3286-engelhardt_accuracy_2016}.

\begin{figure*}
  \includegraphics[width=1\textwidth]{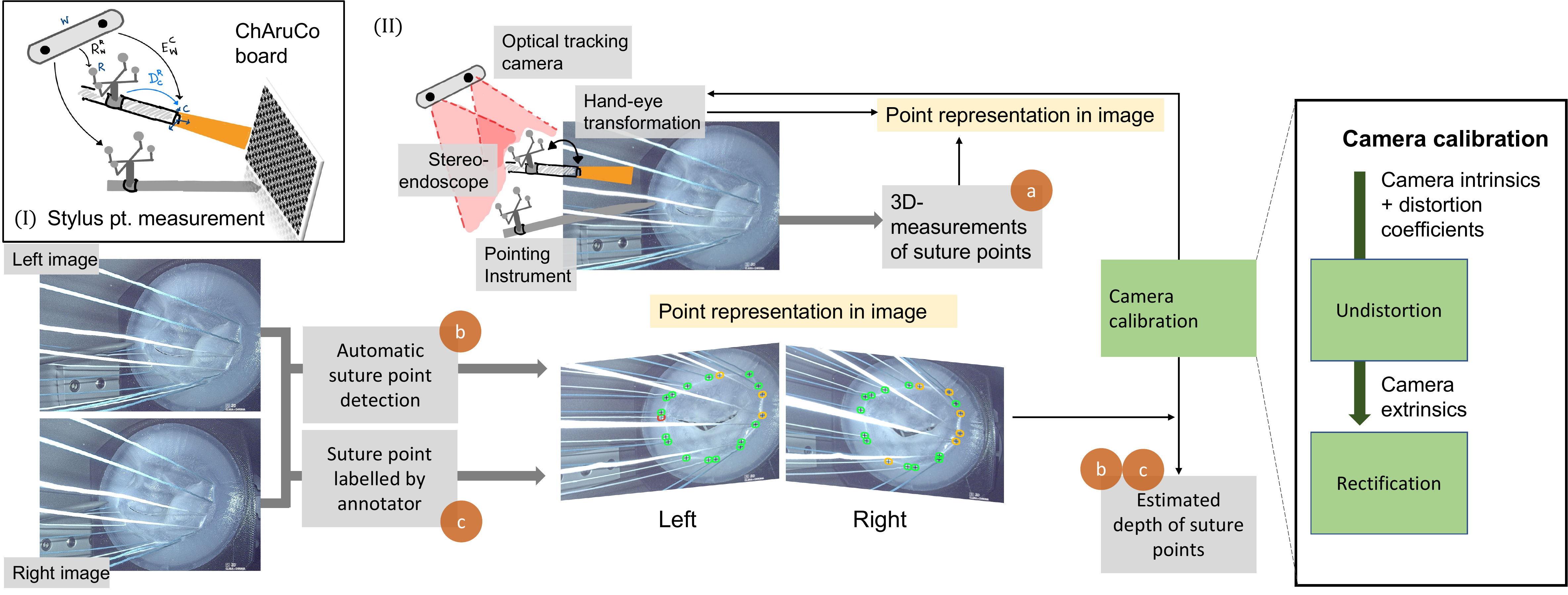}
\caption{Overview of the depth estimation setups that are compared in this work.}
\label{3286-fig:overview}
\end{figure*}
A complementary approach to optical tracking, which does not rely on additional hardware but on the same camera calibration parameters, is to exploit the stereo relation between left and right endoscopic camera frames directly and to detect points which are of interest for distance computation.
In particular, the use of deep learning-based models were previously demonstrated for detecting the entry and exit points of annulus sutures in  simulation and surgery \cite{3286-sharan_point_2021}, using a heatmap based multi-instance point detection approach.

The aim of this work is to compare two different setups for recovering depth to support surgical valve repair. Firstly, we use the previously proposed optical tracking based surgical assistance system \cite{3286-engelhardt_accuracy_2016} together with hand-eye calibration (Fig.\ref{3286-fig:overview} (II)a). Secondly, we use a deep learning based method to detect relevant points from the left and right stereo-endoscopic images \cite{3286-sharan_point_2021}; subsequently, the depth at these points from the image pair is estimated (Fig.\ref{3286-fig:overview} (II)b). Finally, both the methods are compared to manually labeled suture points on the stereo frames, which provides a ground-truth for image-based point detection and a weak ground truth for depth estimation (Fig.\ref{3286-fig:overview} (II)c).

\section{Materials and methods}

\subsection{Dataset}
Mitral valve annuloplasty procedures performed on a surgical simulator with patient-specific valve replica are captured as a video stream in a resolution of $1080\times1920$ at $25 \ts \text{fps}$. 
The \textit{Image1S} stereo-endoscope with $30^{\circ}$ degree optics (Karl Storz SE \& Co. KG, Tuttlingen, Germany) is used.
Two different data sets $data_1$ with $29$ frames and $data_2$ with $32$ frames was selected from the video streams. Furthermore, $12$ frames of a rigid phantom, same as in \cite{3286-engelhardt_accuracy_2016}, are used to test the depth measured with the optical tracking system (NDI Polaris Spectra, Northern Digital Inc., Waterloo, Canada). A mean annulus curve computed over manual segmentations from 42 patients served as a base for the phantom. Twelve approximately equidistant small holes on the surface of the phantom serve as a surrogate for the annuloplasty sutures (Fig. \ref{3286-fig:exampleImage}a,b).

\subsection{Optical tracking and hand-eye calibration}
To obtain depth measurements from the frame of the optical tracking system, referred to here as the world origin $W$, the transformation to the camera frame $C$ is required. To track the relative position of the camera, we attach a fixed reference marker $R$, through which we can determine the unknown transformation $D_R^C$ (Fig. \ref{3286-fig:overview}(I)). To compute this, we first estimate the transformation of the camera $E_W^C$ from the $2D$ image points obtained from images of a calibration pattern, and further measure the corresponding $3D$ points with a stylus tool. The stylus tool is tracked by the optical tracking system, and is calibrated by pivoting the tool around a pin to calculate the offset from the tracked markers. The pivot calibration was performed with $500$ images of the tracking tool. We used a custom precision machined \textit{ChArUco} calibration board (calib.io, Svendborg,
Denmark) to avoid printing and planarity errors, together with the OpenCV calibration library. 

The \textit{EPnP} pose estimation algorithm \cite{3286-Lepetit2008} implementation from OpenCV is then used to calculate the extrinsic camera parameters from the $2D$ and $3D$ points. The transformation between the camera $C$ and the rigid marker $R$ can be calculated with
\begin{equation}\label{3286-eq:rigid2Cam}
    D_R^C = E_W^C \cdot (R_W^R)^{-1}
\end{equation}
A change in the camera position, can be accordingly calculated with the tracked transformation $R_W^R$ 
\begin{equation}\label{3286-eq:camExtrinsics}
    E_W^C =  D_R^C \cdot R_W^R
.\end{equation}
By measuring the image points with a tracked stylus tool, we can compute the transformation from the camera position. The points that are reprojected using this method are denoted by $p_{reproj}$, and the depth measurements are denoted by $d_{he}$. All transformations are computed using a $4x4$ affine transformation matrix with $R|t$.

\subsection{Image-based point detection}
\textit{Ground Truth:} The suture points $p_{gt}$ and point correspondences were manually labeled on the stereo frames by an annotator. Only sutures that can be seen on both the images of the stereo-pair are considered. The labeled points $p_{gt}$ were used for point triangulation to estimate the respective depth $d_{p_{gt}}$ (Fig.\ref{3286-fig:overview}(c)). Note that due to occlusions, some points which could be measured by the tracked stylus could not be labeled in the images.

\textit{Deep Learning:} The deep learning based heatmap multi-instance point detection method was developed in previous work \cite{3286-sharan_point_2021} and is based on the U-net  \cite{3286-ronneberger_u-net_2015}. A Gaussian distribution of $\sigma=2$ was applied to the masks of the training dataset, which has a size of $2800$ images and an image resolution of $288\times512$.
The model takes as input left or right endoscopic images independently from each other and outputs the locations of the suture points as predictions. Then, two filter steps are applied: Points that lie within a threshold of 6 $px$ radius are considered as \textit{true positives}; point correspondence between left and right image were then determined according to the ground truth. Only the points that are present in both the left and the right images are used for further depth computation $d_{p_m}$.
This leads us to a total of 148 point matches from 327 left and 257 right predicted image points in $data_1$ and in $data_2$ is a total of 156 point matches from 275 left and 239 right predicted image points.

\subsection{Evaluation}
\label{3286-subsec:evaluation}

To evaluate the depth values obtained from the two setups, and their respective comparisons with the labeled suture points, we use three kinds of error, namely: $2D$ point error in $px$, $3D$ point error in $mm$ and the distance between the suture points in $3D$ in $mm$.
The Euclidean distance is computed to measure the $2D$ point error for each suture point in the frame. In case of the measured suture point with the optical tracking system, we first project the $3D$ object points to the image coordinate system, and then compute the Euclidean distance to show the $2D$ error per suture point per frame (differences between $p_{reproj}, p_{gt}$, and $p_m$).
Additionally we show the error in $3D$ space from the left endoscopic camera, to get a better understanding of the performance of depth-estimation (differences between $d_{p_{gt}}, d_{p_m}$, and $d_{he}$). Since we are also interested in the correct $3D$ relationship of the suture points, we compare the distance between neighboring suture points to a manually measured distance with a caliper. 
It is to be noted that the errors introduced by camera calibration is the same for all the three methods compared in the following.

\section{Results and Discussion}

\begin{figure*}
  \includegraphics[width=1\textwidth]{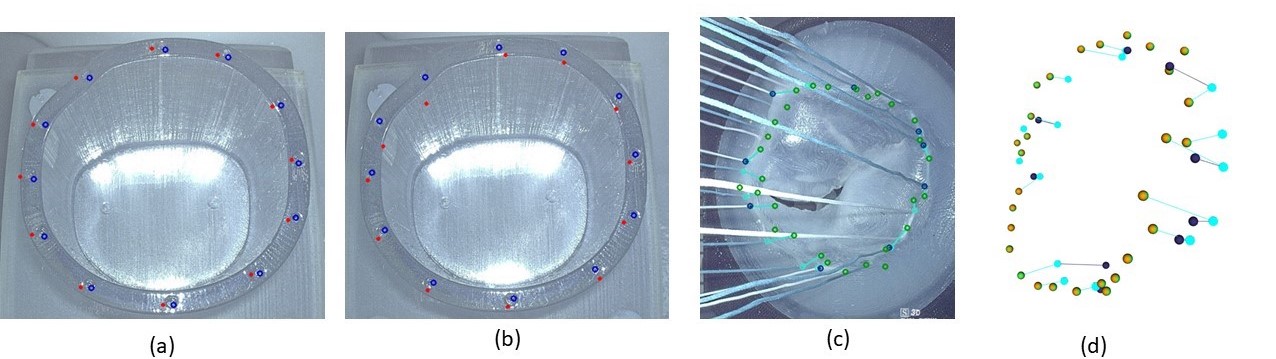}
\caption{The left two images show the reprojected $3D$ points $p_{reproj}$ measured by the stylus tools (red) and manually labeled $2D$ points $p_{gt}$ (blue). 
(a) Measurement with short stylus tool on rigid phantom results in a mean re-projection error of $10.2 \ts \text{px}$; (b) with a long stylus tool a mean re-projection error of $13.73 \ts \text{px}$ to the labeled points is achieved.
(c) 2D image points on the left stereo image and (d) corresponding 3D points with back projection stylus points (green), label points $d_{gt}$ (turquoise), model points $d_m$(black) and 3D stylus measurement $d_{he}$ (orange).}
\label{3286-fig:exampleImage}
\end{figure*}

The error of pivot calibration of the short and long styli is reported in Table \ref{3286-tab:pivot}. 
Furthermore, $12$ points were measured using two different stylus tools, on a mitral valve rigid phantom, to compare the re-projection error (cf. Fig. \ref{3286-fig:exampleImage}). Using the short stylus resulted in a re-projection error of $10.2\ts \text{px}$, and the long stylus in $13.73\ts \text{px}$. For the simulator datasets, the hand-eye calibration is computed on three different camera poses with multiple point sets comprising approximately $50$ points, which result in a re-projection error of $1.49\pm0.44 \ts \text{px}$. The resulting depth computation has a mean deviation of $0.8 \ts \text{mm}$. The calibration with the least error is used for results computation. 
\begin{table}[t]
\caption{Pivot calibration error.}
\label{3286-tab:pivot}
\begin{tabular*}{\textwidth}{l@{\extracolsep\fill}llll}
\hline
                            & Default NDI req. & Short stylus tool  & Long stylus tool\\ \hline
3D RMS Error                & $0.4 \ts \text{mm}$          & $0.36 \ts \text{mm}$          & $0.58 \ts \text{mm}$     \\
Mean Error                  & -                & $0.34 \ts \text{mm}$          & $0.51 \ts \text{mm}$       \\
Maximum 3D Error            & $0.5 \ts \text{mm}$         & $0.48 \ts \text{mm}$          & $1.78 \ts \text{mm}$      \\
Major Angle                 & $45^{\circ}$     & $60.82^{\circ}$    & $55.19^{\circ}$ \\
Minor Angle                 & $45^{\circ}$     & $54.92^{\circ}$    & $47.39^{\circ}$ \\ \hline
\end{tabular*}
\end{table}
Table \ref{3286-tab:2DError} shows the error in $2D$ space, for the left and right images, for each of the three methods that were used to triangulate the points and compute the depth. The error was computed over $26$ and $30$ image points. In comparison with the ground-truth points, the model points $p_m$ produced a lower error of $10.91 \ts \text{px}$ compared to the reprojected points $p_{reproj}$ (cf. Table \ref{3286-tab:2DError}), while the error between the model points and the reprojected points was the highest with $13.86 \ts \text{px}$.
In the $3D$ error computation (cf. Table \ref{3286-tab:3DError}), the error between the ground truth points and the points measure from the stylus from $data_2$ was the highest with $4.21\pm2.86\ts \text{mm}$, and the error between the predicted points $d_{p_m}$ from the model and the labelled points $d_{p_{gt}}$ from $data_1$ was the least with $1.47\pm2.08 \ts \text{mm}$. The depth between the measured points $d_{he}$ and the depth from the reprojected points are similar, with a difference of $0.0035\pm0.00745 \ts \text{mm}$ and $0.0074\pm0.015 \ts \text{mm}$ between each other.

\begin{table}[t]
\centering
\caption{$2D$ error computed for the suture points in $px$.}
\label{3286-tab:2DError}
\begin{tabular*}{\textwidth}{l@{\extracolsep\fill}lllllll}
\hline
                & \multicolumn{2}{c}{$p_{gt}-p_{reproj}$} & \multicolumn{2}{c}{$p_{m}-p_{reproj}$}  & \multicolumn{2}{c}{$p_{m}-p_{gt}$}\\
                \hline
                & Left            & Right                   & Left             & Right           & Left             & Right           \\
$data_1$        & $11.74\pm7.98$    & $11.55\pm7.63$        & $11.23\pm7.07$   & $10.91\pm7.16$  & $2.09\pm1.06$    & $ 2.2\pm1.19$   \\
$data_2$        & $10.64\pm8.29$    & $10.22\pm8.2$         & $13.86\pm9.72$   & $13.82\pm9.8$   &$2.29\pm1.32$     & $2.37\pm1.27$ \\ \hline
\end{tabular*}
\end{table}

\begin{table}[t]
\caption{$3D$ error computed for the suture points in $mm$.}
\label{3286-tab:3DError}
\begin{tabular*}{\textwidth}{l@{\extracolsep\fill}llll}
\hline
                            & $d_{p_{gt}}-d_{he}$    & $d_{p_m}-d_{p_{gt}}$    & $d_{p_m}-d_{he}$ \\\hline
$data_1$                    & $3.77\pm2.83$     & $1.47\pm2.08$     & $4.14\pm5.45$ \\
$data_2$                    & $2.97\pm1.89$     & $2.01\pm2.79$     & $3.06\pm2.74$ \\
\hline
\end{tabular*}  
\end{table}

To sum up, when compared to the labelled points, the error measured in $2D$ and $3D$ space is least for the points predicted from the deep learning model in comparison with the optical tracking system. Compared to distances measured with a caliper, the error in $d_{he}$ is the highest with $2.1\pm1.28\ts \text{mm}$, while $d_{p_m}$ performed the best with an error of $0.46\pm0.44\ts \text{mm}$ and $d_{p_{gt}}$ with an error of $0.7\pm0.62 \ts \text{mm}$. However, note in this work, only the \textit{True Positive} predictions from the model are taken into consideration and those which have a corresponding point in the other image. Automatic matching was not considered so far.
In conclusion, our preliminary results demonstrate the advantages in accuracy when considering the correctly identified points in endoscope-based image point reconstruction over optically tracked stylus measurements. 

\begin{acknowledgement}
The research was supported by Informatics for Life project funded by the Klaus Tschira Foundation and the DFG through grant INST 35/1314-1 FUGG, INST 35/1503-1 FUGG,  DE  2131/2-1,  and  EN  1197/2-1.
\end{acknowledgement}

\printbibliography

\end{document}